# GaKCo: a Fast <u>Ga</u>pped *k*-mer string <u>K</u>ernel using <u>Co</u>unting

Ritambhara Singh, Arshdeep Sekhon, Jack Lanchantin,
Kamran Kowsari, Beilun Wang and Yanjun Qi

Department of Computer Science, University of Virginia (`yanjun@virginia.edu`)

**Abstract.** String Kernel (SK) techniques, especially those using gapped *k*-mers as features (gk), have obtained great success in classifying sequences like DNA, protein, and text. However, the state-of-the-art gk-SK runs extremely slow when we increase the dictionary size ($\Sigma$) or allow more mismatches ($M$). This is because current gk-SK uses a trie-based algorithm to calculate co-occurrence of mismatched substrings resulting in a time cost proportional to $O(\Sigma^M)$. We propose a **fast** algorithm for calculating <u>Ga</u>pped *k*-mer <u>K</u>ernel using <u>Co</u>unting (GaKCo). GaKCo uses associative arrays to calculate the co-occurrence of substrings using cumulative counting. This algorithm is fast, scalable to larger $\Sigma$ and $M$, and naturally parallelizable. We provide a rigorous asymptotic analysis that compares GaKCo with the state-of-the-art gk-SK. Theoretically, the time cost of GaKCo is independent of the $\Sigma^M$ term that slows down the trie-based approach. Experimentally, we observe that GaKCo achieves the same accuracy as the state-of-the-art and outperforms its speed by factors of 2, 100, and 4, on classifying sequences of DNA (5 datasets), protein (12 datasets), and character-based English text (2 datasets). [1]

## 1 Introduction

Sequence classification is one of the most important machine learning tasks, with widespread uses in fields like biology and natural language processing. Besides accuracy, speed is a critical requirement for modern sequence classification methods. For example, with the advancement of sequencing technologies, a massive amount of protein and DNA sequence data is produced daily [22]. There is an urgent need to analyze these sequences quickly for assisting time-sensitive experiments. Similarly, on-line information retrieval systems need to classify text sequences, for instance when quickly assessing customer reviews or categorizing documents to different topics.

In this paper, we focus on the String Kernels (SK) in the Support Vector Machine (SVM) framework for supervised sequence classification. SK-SVM methods have been successfully used for classifying sequences like DNA [18, 15, 3, 27], protein [13] or character based natural language text [28]. They have provided state-of-the-art classification accuracy and can guarantee nice asymptotic behavior due to SVM's convex formulation and theoretical property [31]. Through comparing length-*k* local substrings (*k*-mers) and incorporating mismatches and gaps, this category of models calculates the similarity (i.e., so-called kernel function) among sequence samples. Then, using such similarity measures, SVM is trained to classify sequences. Recently, Ghandi et al. [11] developed the state-of-the-art SK-SVM tool called gkm-SVM. gkm-SVM uses a gapped *k*-mer formulation [12] that reduces the feature space considerably compared to other *k*-mer based SK approaches.

Existing *k*-mer based SK methods can become very slow or even unfeasible when we increase (1) the number of allowed mismatches ($M$) or (2) the size of the dictionary ($\Sigma$) (detailed asymptotic analysis in Section 2). Allowing mismatches during substring comparisons is important since most sequences in biology are prone to mutations, i.e., insertions, deletions or substitution of sequence characters. Also, the size of the dictionary varies from one sequence classification domain to another. While DNA sequence is composed of only four characters ($\Sigma = 4$), most other domains have bigger dictionary sizes like for proteins, $\Sigma = 20$ and for character-based English text, $\Sigma = 36$. The state-of-the-art tool, gkm-SVM, may work well for cases with small values of $\Sigma$ and $M$ (like for DNA sequences with $\Sigma = 4$ and $M < 4$), however, its kernel calculation is slow for cases like DNA with larger $M$, protein (dictionary size = 20), or character-based English text sequences (dictionary size = 36). Its trie-based implementation, in the worst case, scales exponentially with the dictionary size and the number of mismatches ($O(\Sigma^M)$). For example, gkm-SVM takes more than 5 hours to calculate the kernel matrix for one protein sequence classification task with only 3312 sequences. This speed limitation hinders the practical applications of SK-SVM.

---

[1] GaKCo is shared as an open source tool at `https://github.com/QData/GaKCo-SVM`

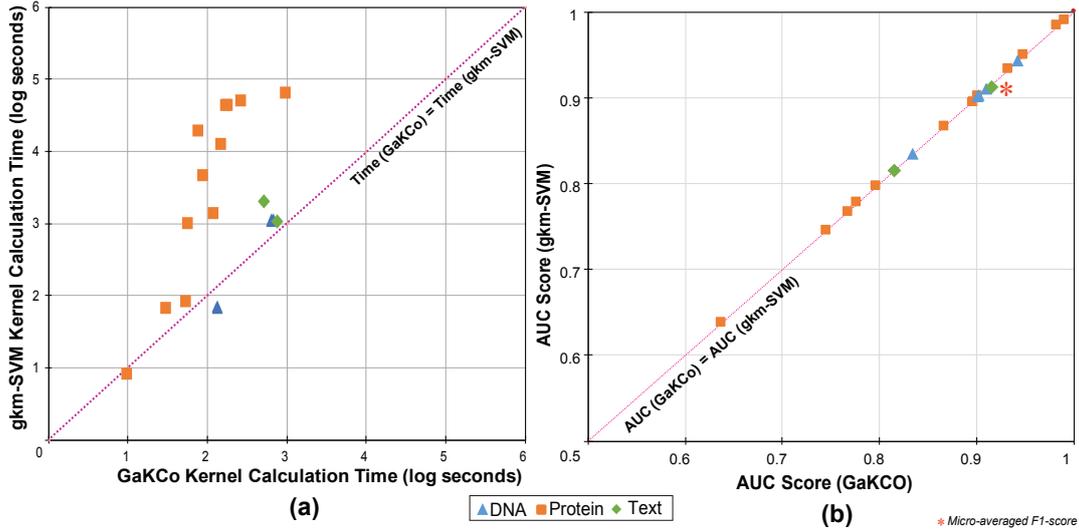

**Fig. 1.** (a) Kernel calculation times (log(seconds)) of GaKCo (X-axis) versus gkm-SVM (Y-axis) for 19 different datasets - protein (12), DNA (5), and text (2). GaKCo is faster than gkm-SVM for 16/19 datasets. (b) Empirical performance for the same 19 datasets (DNA, protein, and text) of GaKCo (X-axis) versus gkm-SVM (Y-axis). GaKCo achieves the same AUC-scores as gkm-SVM.

This paper proposes a **fast** algorithmic strategy, GaKCo: **Ga**pped $k$-mer **K**ernel using **Co**unting to speed up the gapped $k$-mer kernel calculation. GaKCo uses a "sort and count" approach to calculate kernel similarity through cumulative $k$-mer counting [15]. GaKCo groups the counting of co-occurrence of substrings at each fixed number of mismatches ($\{0, \ldots, M\}$) into an independent procedure. Such grouping significantly reduces the number of updates on the kernel matrix (an operation that dominates the time cost). This algorithm is naturally parallelizable; therefore we present a multithread variation as our ultimate tool that improves the calculation speed even further.

We provide a rigorous theoretical analysis showing that GaKCo has a better asymptotic kernel computation time cost than gkm-SVM. Our empirical experiments, on three different real-world sequence classification domains, validate our theoretical analysis. For example, for the protein classification task mentioned above where gkm-SVM took more than 5 hours, GaKCo takes only 4 minutes. Compared to GaKCo, gkm-SVM slows down considerably especially when $M \geq 4$ and for tasks with $\Sigma \geq 4$. Experimentally, GaKCo provides a speedup by factors of 2, 100 and 4 for sequence classification on DNA (5 datasets), protein (12 datasets) and text (2 datasets), respectively, while achieving the same accuracy as gkm-SVM. Fig. 1(a) compares the kernel calculation times of GaKCo (X-axis) with gkm-SVM (Y-axis). We plot the kernel calculation times for the best performing $(g, k)$ parameters (see supplementary GitHub) for 19 different datasets. We see that GaKCo is faster than gkm-SVM for 16 out of 19 datasets that we have tested. Similarly, we plot the empirical performance (AUC scores or F1-score) of GaKCo (horizontal axis) versus gkm-SVM (vertical axis) for the best performing $(g, k)$ parameters (see supplementary) for the 19 different datasets in Fig. 1(b). It shows that the empirical performance of GaKCo is same as gkm-SVM with respect to the AUC scores. In summary, the main contributions of this work are:

– **Fast:** GaKCo is a novel combination of two efficient concepts: (1) reduced gapped $k$-mer feature space and (2) associative array based counting method, making it faster than the state-of-the-art gapped $k$-mer string kernel, while achieving the same accuracy.
– GaKCo can **scale up** to larger values of $m$ and $\Sigma$.
– **Parallelizable:** GaKCo algorithm lends itself to a naturally parallelizable implementation.
– We also present a detailed **theoretical analysis** of the asymptotic time complexity of GaKCo versus state-of-the-art gkm-SVM. This analysis, to our knowledge, has not been reported before.

Table 1. List of symbols and their descriptions that are used.

| Notations | Descriptions |
|---|---|
| $D$ | Dataset under consideration, $D = \{x_1, x_2, \ldots, x_N\}$ |
| $N$ | Number of sequences in a given dataset $D$ |
| $x, x'$ | Pair of strings in $D$ that are compared for kernel calculation |
| $K(x, x')$ | Kernel Function; Eq. (7) is for the gapped k-mer case |
| $\phi(x)$ | Feature space representation of the string x |
| $l$ | Average length of sequences in a given dataset $D$ |
| $\Sigma$ | Size of the dictionary of a given dataset $D$ |
| $g$ | Length of the gapped instance or $g$-mer (specified by the user) |
| $k$ | Length of $k$-mer inside a gapped instance (specified by the user) |
| $M$ | $M = (g - k)$; maximum number of mismatches allowed between two $g$-mers; |
| $m$ | Number of mismatches between two $g$-mers. $m \in \{0, \ldots M\}$ |
| $c_{gk}$ | $c_{gk} = \sum_{m=0}^{M=(g-k)} \binom{g}{m}$. |
| $u$ | Number of unique $g$-mers in a given dataset $D$ |
| $z$ | Number of unique $g$-mers with $> 1$ occurrence in a given dataset $D$ |
| $\mathbf{N}_m(x, x')$ | Mismatch profile: number of matching $g$-mer pairs between $x$ and $x'$ when allowing $m$ mismatches; see Eq. (9) |
| $\mathbf{C}_m(x, x')$ | Cumulative mismatch profile: number of matching $\{g - m\}$-mer pairs between $x$ and $x'$. Each $\{g-m\}$-mer is generated from a $g$-mer by removing characters from a total of $m$ different positions; See Eq. (8) |
| $\eta$ | Average size of the *nodelist* of leafnodes in gkm-SVM's trie. Each leafnode is a unique $g$-mer whose *nodelist* includes all $g$-mers in the trie whose hamming distance to this leaf is up to $M$; See Eq. (10) |

The rest of the paper is organized as follows: Section 2 introduces the details of GaKCo and theoretically proves that asymptotically GaKCo runs faster than gkm-SVM for a large dictionary or allowing for more mismatches. Then Section 3 provides the experimental results we obtain on three major benchmark applications: TFBS binding prediction (DNA), Remote Protein Homology prediction (Proteins) and Text Classification (categorization and sentiment analysis). Empirically, GaKCo shows consistent improvements over gkm-SVM in computation speed across different types of datasets. When allowing a higher number of mismatches, the disparity in speed between GaKCo and the baseline becomes more apparent. Table 1 summarizes the important notations we use. Due to the space limitation, we discuss the related studies in the supplementary. Recently, Deep Neural Networks (NNs) have provided state-of-the-art performances for various sequence classification tasks. We compare GaKCo's empirical performance with a state-of-the-art deep convolutional neural network (CNN) model [17]. On datasets with few training samples, GaKCo achieves an average accuracy improvement of 20% over the CNN model (details in the supplementary).

## 2 Method

### 2.1 Background: Gapped *k*-mer String Kernels

The key idea of string kernels is to apply a function $\phi(\cdot)$, which maps strings of arbitrary length into a vectorial feature space of fixed dimension. In this space, we apply a standard classifier such as Support Vector Machine (SVM) [31]. Kernel versions of SVMs calculate the decision function for an input $x$ as:

$$f(x) = \sum_{i=1}^{N} \alpha_i y_i K(x_i, x) + b \qquad (1)$$

where $N$ is the total number of training samples and $K(\cdot, \cdot)$ is a *kernel function*. String kernels ([18, 15, 11]), implicitly compute $K(x, x')$ as an inner product in the feature space:

$$K(x, x') = \langle \phi(x), \phi(x') \rangle, \qquad (2)$$

where $x = (s_1, \ldots, s_{|x|})$. $x, x' \in \mathcal{S}$. $|x|$ denotes the length of the string $x$. $\mathcal{S}$ represents the set of all strings composed from a dictionary $\Sigma$. The mapping $\phi : \mathcal{S} \to \mathbb{R}^p$ takes a sequence $x \in \mathcal{S}$ to a $p$-dimensional feature vector.

The feature representation $\phi(\cdot)$ plays a vital role in string analysis since it is hard to describe strings as feature vectors. One classical method is to represent it as an unordered set of $k$-mers, or combinations of $k$ adjacent characters. A feature vector indexed by all $k$-mers records the number of occurrences of each $k$-mer in the current string. The string kernel using this representation is called spectrum kernel [19], where the spectrum representation counts the occurrences of each $k$-mer in a string. Kernel scores between strings are computed by taking an inner product between corresponding "$k$-mer-indexed" feature vectors:

$$K(x, x') = \sum_{\gamma \in \Gamma_k} c_x(\gamma) \cdot c_{x'}(\gamma) \qquad (3)$$

where $\gamma$ represents a $k$-mer, $\Gamma_k$ is the set of all possible $k$-mers, and $c_x(\gamma)$ is the number of occurrences (normalized) of $k$-mer $\gamma$ in string $x$. Many variations of spectrum kernels ([15, 32, 14, 8]) exist in the literature that mostly extend it by including mismatched $k$-mers when calculating the number of occurrences.

Spectrum kernel and its mismatch variations generate extremely sparse feature vectors for even moderately sized values of $k$, since the size of $\Gamma_k$ is $\Sigma^k$. To solve this issue, Ghandi et al. [12] introduced a new set of feature representations, called *gapped $k$-mers*. It is characterized by two parameters: (1) $g$, the size of a substring with gaps (we call this gapped instance as $g$-mer hereafter) and (2) $k$, the size of non-gapped substring in a $g$-mer (we call it $k$-mer). The number of gaps is $(g-k)$. The inner product to compute the gapped $k$-mer kernel function includes sum over all possible $k$-mer feature counts obtained from the $g$-mers:

$$K(x, x') = \sum_{\gamma \in \Theta_g} c_x(\gamma) \cdot c_{x'}(\gamma) \qquad (4)$$

where $\gamma$ represents a $k$-mer, $\Theta_g$ is the set of all possible gapped $k$-mers that can appear in all the $g$-mers (each with $(g-k)$ gaps) in a given dataset (denoted as $D$ hereafter) of sequence samples.

This formulation's advantage is that it drastically reduces the number of $k$-mers to consider. If we sum over all $k$-mers, as in Eq. (3), each of the $\binom{g}{k}$ "non-gap" positions in the $g$-mer may be filled with any of $\Sigma$ letters. Thus, the sum has $\binom{g}{k}\Sigma^k$ terms — the number of possible gapped $k$-mers. This feature space grows rapidly with both $\Sigma$ and $k$. In contrast, Eq. (4) (implemented as gkm-SVM [11]) includes only those $k$-mers whose gapped formulation has appeared in the dataset, $D$. $\Theta_g$ includes all unique $g$-mers of the dataset $D$, whose size $|\Theta_g|$ is normally much smaller than $\binom{g}{k}\Sigma^k$ because the new feature space is restricted to only observable gapped $k$-mers in $D$. Ghandi et al. [11] use this intuition to reformulate Eq. (4) into:

$$K(x, x') = \sum_{i=0}^{l_1} \sum_{j=0}^{l_2} h_{gk}(g_i^x, g_j^{x'}) \qquad (5)$$

For two sequences $x$ and $x'$ of lengths $l_1$ and $l_2$ respectively. $g_i^x$ and $g_j^{x'}$ are the $i^{th}$ and $j^{th}$ $g$-mers of sequences $x$ and $x'$ (i.e., $g_i^x$ is a continuous substring of $x$ starting from the $i$-th position and ending at the $(i + g - 1)^{th}$ position of $x$). $h_{gk}$ represents the inner product (or similarity) between $g_i^x$ and $g_j^{x'}$ using the co-occurrence of gapped k-mers as features. $h_{gk}(g_i^x, g_j^{x'})$ is non-zero only when $g_i^x$ and $g_j^{x'}$ have common $k$-mers.

**Definition 1.** $\mathbf{g\text{-}pair}_m(x, x')$ *denotes a pair of $g$-mers $(g_1^x, g_2^{x'})$ whose Hamming distance is exactly $m$. $g_1^x$ is from sequence $x$ and $g_2^{x'}$ is from sequence $x'$.*

Each $\mathbf{g\text{-}pair}_m(.)$ has $\binom{g-m}{k}$ common $k$-mers, therefore its $h_{gk}$ can be directly calculated as $h_{gk}(\mathbf{g\text{-}pair}_m) = \binom{g-m}{k}$. Ghandi et al. [11] formulate this observation formally into the coefficient $h_m$:

$$h_m = \begin{cases} \binom{g-m}{k}, & \text{if } g - m \geq k \\ 0, & \text{otherwise.} \end{cases} \qquad (6)$$

$h_m$ describes the co-occurrence count of common $k$-mers for each possible $\mathbf{g\text{-}pair}_m(.)$ in $D$. $h_m > 0$ only for cases of $m \leq (g-k)$ or $(g-m) \geq k$. This is because there will be no common $k$-mers when the number of mismatches ($m$) between two $g$-mers is more than $(g-k)$. Now we can reformulate Eq. 5 by grouping $\mathbf{g\text{-}pairs}_m(x, x')$ with respect to different values of $m$. This is because $\mathbf{g\text{-}pairs}_m(.)$ with same $m$ contribute the same number of co-occurrence counts: $h_m$. Thus, Eq. 5 can be adapted into the following compact form:

$$K(x, x') = \sum_{m=0}^{g-k} N_m(x, x') h_m \qquad (7)$$

$N_m(x, x')$ represents the number of $\mathbf{g\text{-}pair}_m(x, x')$ between sequence $x$ and $x'$. $N_m(x, x')$ is named as *mismatch profile* by [11]. Now, to compute kernel function $K(x, x')$ for gapped $k$-mer SK, we only need to calculate $N_m(x, x')$ for $m \in \{0, \ldots g-k\}$, since $h_m$ can be precomputed [2]. The state-of-the-art tool gkm-SVM [11] calculates $N_m(x, x')$ using a trie based data structure that is similar to [18] (with some modifications, details in Section 2.3).

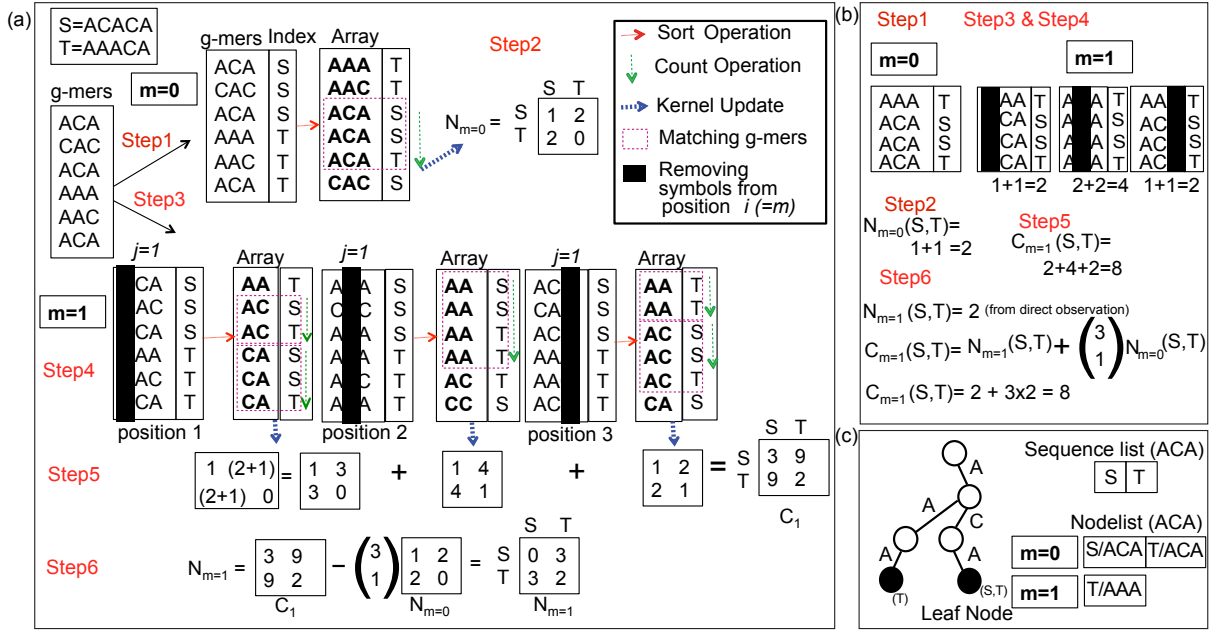

**Fig. 2.** (a) Overview of GaKCo algorithm for calculating mismatch profile $N_m(S,T)$, where $S = ACACA$ and $T = AAACA$, and $g = 3$ forming g-mers $\{ACA, CAC, ACA\}$ and $\{AAA, AAC, ACA\}$ respectively. [Step 1] For $m = 0$, all $g$-mers are sorted lexicographically. [Step 2] $N_{m=0}(S,T)$ is calculated directly by sorting and counting. [Step 3] For $m = 1$, we perform over counting of the $g-1$-mers by picking 1 position at a time (from $\binom{g=3}{1}$ positions) and removing symbols to obtain $(g-1)$-mers. [Step 4] We sort and count to find the number of matching $(g-1)$-mers for each picked position. [Step 5] Summing up over all $\binom{g=3}{1}$ positions, we get *cumulative mismatch profile* $\mathbf{C}_{m=1}$. [Step 6] Using Eq. 9 we get $N_{m=1}(S,T) = 3$ from $C_{m=1}(S,T) = 9$ and $N_{m=0}(S,T) = 2$. This count is equal to the actual number of pairs of g-mers at Hamming distance $m = 1$ between $s$ and $t$ (i.e. $\{ACA : s/2, AAA : t/1\}, \{CAC : s/1, AAC : t/1\}$). A case demonstration of (b) the overcounting when calculating $\mathbf{C}_{m=1}$ (c) two leafnode g-mers and associated nodelist for leaf $\{ACA\}$ in the trie used by gkm-SVM.

### 2.2 Proposed: Gapped $k$-mer Kernel with Counting (GaKCo)

In this paper, we propose GaKCo, a fast and novel algorithm for calculating gapped $k$-mer string kernel. GaKCo provides superior time performance over the state-of-the-art gkm-SVM and is different from it in three aspects:

- **Data Structure.** gkm-SVM uses a trie based data structure (plus a separate nodelist at each leafnode) for calculating $N_m$ (see Figure 2(c)). In contrast, GaKCo uses simple arrays with a "sort-and-count" approach.
- **Algorithm.** GaKCo performs $g$-mer based cumulative counting of co-occurrence to calculate $N_m$.
- **Parallelization.** GaKCo groups computations for each value of $m$ into an independent function, making it naturally parallelizable. We, therefore, provide a parallel version that uses multithread implementation.

***Intuition*** : When calculating $\mathbf{N}_m$ between all pairs of sequences in $D$ for each value of $m$ ($m \in \{0, \ldots, M = g-k\}$), we can use counting to process all **g-pairs**$_m(.)$ (details below) from $D$ together. Then we can calculate $\mathbf{N}_m$ from such count statistics of **g-pairs**$_m(.)$. This method is entirely different from gkm-SVM that uses a trie to organize g-mers such that each leafnode's (a unique g-mer's) nodelist points to its mismatched g-mer neighbors in $D$.

---
[2] For convenience, we will occasionally identify the map $N_m(\cdot, \cdot)$ with the $N \times N$ matrix $\mathbf{N}_m$, consisting of the application of $N_m$ to each pair of sequences $x, x' \in D$. This convention is also followed for the kernel function, $K(\cdot, \cdot) \to \mathbf{K}$, and the cumulative mismatch profile (introduced later), $C_m(\cdot, \cdot) \to \mathbf{C}_m$.

***Algorithm*** GaKCo calculates $N_m(x, x')$ as follows (pseudo code: Algorithm 1):

1. GaKCo first extracts all possible $g$-mers from all the sequences in $D$ and puts them in a simple array. Given that there are $N$ number of sequences with average length $l$ [3], the total number of $g$-mers is $N \times (l - g + 1) \sim Nl$ (see Fig. 2 (a)).

2. $N_{m=0}(x, x')$ represents the number of **g-pair**$_{m=0}(x, x')$ (pairs of $g$-mers whose Hamming distance is 0) between $x$ and $x'$. To compute $N_{m=0}(x_i, x_j) \ \forall i, \forall j = 1, ..., N$, GaKCo sorts all the $g$-mers lexicographically (see Fig. 2(a) [Step 1]) and counts the occurrences (if $> 1$) of each unique $g$-mer. Then we use these counts and the associated indexes of sequences to update all the kernel entries for sequences that include the matching $g$-mers (Fig. 2(a) [Step 2]). This computation is straight-forward and the sort and count step takes $O(gNl)$ time cost while the kernel update costs $O(zN^2)$ (at the worst case). Here, $z$ is the number of $g$-mers that occur $> 1$ times.

3. For cases when $m = 1, \ldots (g-k)$, we use a statistics measure $C_m(x, x')$, called *cumulative mismatch profile* between two sequences $x$ and $x'$. This measure describes the number of matching $(g-m)$-mers between $x$ and $x'$. Each $(g-m)$-mer is generated from a $g$-mer by removing a total number of $m$ positions. We can calculate the exact *mismatch profile* $\mathbf{N}_m$ from the cumulative mismatch profile $\mathbf{C}_m$ for $m > 0$ (see Step 4).

   By sorting the lists of $g$-mers with $m$ ignored entries, we compute $\mathbf{C}_m$. First, we first pick $m$ positions and remove the symbols in those positions from all observed $g$-mers, generating a list of $(g-m)$-mers (Fig. 2 (a) [Step 3]). We then sort and count this list to get the number of matching $(g-m)$-mers (Fig. 2 (b) [Step 4]). For the sequences that have matching $(g-m)$-mers, we add the counts into their corresponding entries in matrix $\mathbf{C}_m$. This sequence of operations is repeated for all $\binom{g}{m}$ selections of $m$ positions. Then, $\mathbf{C}_m$ is equal to the sum of counts from all $\binom{g}{m}$ runs (Fig. 2 [Step 5]).

4. We compute $\mathbf{N}_m$ using $\mathbf{C}_m$ and $\mathbf{N}_j$ for $j = 0, \ldots, m-1$.

   Given two $g$-mers $g_1$ and $g_2$, we remove symbols from the same set of $m$ positions of both $g$-mers to get two $(g-m)$-mers: $g_1'$ and $g_2'$. If the Hamming distance between $g_1'$ and $g_2'$ is zero, then we can conclude that the Hamming distance between the original two $g$-mers is less than or equal to $m$ (formal proof in supplementary). For instance, $C_{m=1}(x, x')$ records the statistic of matching $(g-1)$-mers between $x$ and $x'$. It includes the matching statistics of all $g$-mer pairs with Hamming distance exactly 1, but it also over-counts the matching statistics of all $g$-mer pairs with Hamming distance 0. This is because the matching $g$-mers for $m = 0$ also match for $m = 1$ and contribute to the matching statistics $\binom{g}{1}$ times! This over-counting occurs for other values of $m$ as well. Therefore we can calcualte the cumulative mismatch profile $\mathbf{C}_m$ as: $\forall m \in \{0, \ldots, g-k\}$

$$\mathbf{C}_m = \mathbf{N}_m + \sum_{j=0}^{m-1} \binom{g-j}{m-j} \mathbf{N}_j \quad (8)$$

   We demonstrate this over-counting in Fig. 2(b). Rearranging Eq. 8, we get the exact mismatch profile $N_m$ as:

$$\mathbf{N}_m = \mathbf{C}_m - \sum_{j=0}^{m-1} \binom{g-j}{m-j} \mathbf{N}_j \quad (9)$$

   We subtract $\mathbf{N}_j$ from $\mathbf{C}_m$ to compensate for the over-counting described above.

***Parallelization:*** For each value of $m$ from $\{0, \ldots M = g-k\}$, calculating $\mathbf{C}_m$ is independent from other values of $m$. Therefore, GaKCo's algorithm can be easily revised into a parallel version. Essentially, we just need to revise Step 9 in Algorithm 1 (pseudo code) — "For each value of $m$" — into, "For each value of $m$ per machine/per core/per thread". In our current implementation, we create a thread for each value of $m$ from $\{0, \ldots M = g-k\}$ and calculate $\mathbf{C}_m$ in parallel. In the end, we compute the final kernel matrix $K$ using all the resulting $\mathbf{C}_m$ matrices. Fig. 4 show the improvement of kernel calculation speed of the multi-thread version over the single-thread implementation of GaKCo.

### 2.3 Theoretical Comparison of Time Complexity

In this section, we conduct asymptotic analysis to compare the time complexities of GaKCo with the state-of-the-art toolbox gkm-SVM.

---
[3] A simplification of real world datasets in which sequence length varies across samples

**Table 2.** Comparing time complexity. gvm-SVM's time cost is $O(gNl + \eta ug + \eta uN^2)$. GaKCo's time complexity is $O(c_{gk}[gNl + zN^2])$. In gkm-SVM the $\eta uN^2$ dominates, while the $c_{gk}zN^2$ term dominates for GaKCo.

|  | GaKCo | gkm-SVM |
|---|---|---|
| Pre-processing | $c_{gk}gNl$ | $gNl + \eta ug$ |
| Kernel updates | $c_{gk}zN^2$ | $\eta uN^2$ |

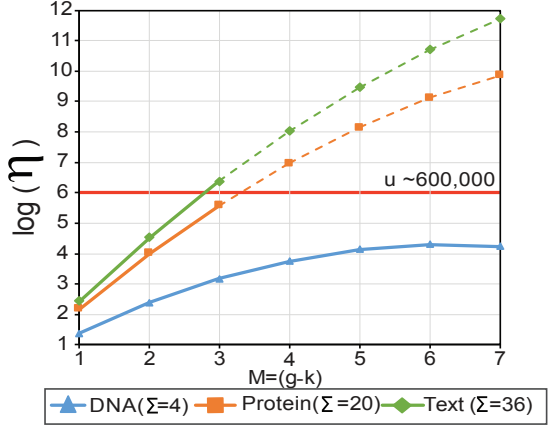

**Fig. 3.** With a growing $g - k$, the growth curve of $\eta$ (Eq. (10): the estimated *nodelist* size in gkm-SVM). Both arguments of the min function are plotted; $\eta$ grows exponentially until $c_{gk}(\Sigma - 1)^M$ exceeds the number of unique $g$-mers, $u$.

**Algorithm 1** GaKCo

**Require:** $L, g, k$ ▷ L=Array list of $g$-mers
1: **procedure** CALCULATEKERNEL($L,g,k$)
2:     $M \leftarrow g - k$
3:     $\mathbf{N} \leftarrow$ MISMATCHPROFILE($L,g,M$)
4:     $K \leftarrow 0$
5:     **for** $m : 0 \to M$ **do**
6:        $h_m \leftarrow \binom{g-m}{k}$
7:        $K \leftarrow K + \mathbf{N}_m \cdot h_m$
8: **procedure** MISMATCHPROFILE($L,g,M$)
9:     **for** $m : 0 \to M$ **do** ▷ Parallel threads
10:       $\mathbf{C}_m \leftarrow 0$ ▷ Cumulative Profile
11:       $n_{pos} \leftarrow \binom{g}{m}$ ▷ Number of positions
12:       **for** $i : 0 \to n_{pos}$ **do**
13:          $\mathbf{C}_m^i \leftarrow 0$
14:          $L^i \leftarrow removePosition(L, i)$
15:          $L^i \leftarrow sort(L^i)$
16:          $\mathbf{C}_m^i \leftarrow countAndUpdate(L^i)$
17:          $\mathbf{C}_m \leftarrow \mathbf{C}_m + \mathbf{C}_m^i$
18:     **for** $m : 0 \to M$ **do**
19:        $\mathbf{N}_m \leftarrow \mathbf{C}_m$
20:        **for** $j : 0 \to m - 1$ **do**
21:           $\mathbf{N}_m \leftarrow \mathbf{N}_m - \binom{g-j}{m-j}\mathbf{N}_j$
     **return N** ▷ $\mathbf{N} = [N_0, \ldots, N_M]$
**Ensure: K** ▷ Kernel Matrix

***Time Complexity of GaKCo:*** The time cost of GaKCo splits into two groups: (1) Pre-processing: those operations that indirectly update the matching statistics among sequences; (2) Kernel updates: those operations that directly update the matching statistics among sequences.

*Pre-processing:* For each possible $m$ ($m \in \{0, \ldots M = g - k\}$), GaKCo needs to choose $m$ positions for symbol removing (Fig. 2 (a) [Step 3]), and then sort and count the possible $(g - m)$-mers from $D$ (Fig. 2 (a) [Step 4]). Therefore the time cost of pre-processing is $O(\Sigma_{m=0}^{M=g-k} \binom{g}{m}(g-m)Nl) \sim O(\Sigma_{m=0}^{M} \binom{g}{m}gNl)$. To simplifying notations, we use $c_{gk}$ to represent $c_{gk} = \Sigma_{m=0}^{M=(g-k)} \binom{g}{m}$ hereafter.

*Kernel Updates:* These operations update the entries of $\mathbf{C}_m$ or $\mathbf{N}_m$ matrices when GaKCo finishes each round of counting the number of matching $(g - m)$-mers. Assuming $z$ denotes the number of unique $(g - m)$-mers that occur $> 1$ times, the time cost of kernel update operations is (at the worst case) equivalent to $O(\Sigma_{m=0}^{M} \binom{g}{m}zN^2) \sim O(c_{gk}zN^2)$. Therefore, the overall time complexity of GaKCo is $O(c_{gk}[gNl + zN^2])$.

***gkm-SVM Algorithm:*** Now we introduce the algorithm of gkm-SVM briefly. Given that there are $N$ sequences in a dataset $D$, gkm-SVM first constructs a trie recording all the unique $g$-mers in $D$. Each leafnode in the trie stores a unique $g$-mer (more precisely by its path to the rootnode) of $D$. We use $u$ to denote the total number of the unique $g$-mers in this trie. Next, gkm-SVM traverses the tree in a depth-first ordering. For each leafnode (like $ACA$ in (Fig. 2 (c)), it maintains a *nodelist* that includes all those $g$-mers in $D$ whose Hamming distance to the leafnode $g$-mer $\leq M$. When accessing a leafnode, all mismatch profile matrices $N_m(x, x')$ for $m \in \{0, \ldots, M = (g - k)\}$ are updated for all possible pairs of sequences $x$ and $x'$. Here $x$ consists of the $g$-mer of the current leafnode (like $S/ACA$ in (Fig. 2 (c)). $x'$ belongs to the *nodelist*'s sequence list. $x'$ includes a $g$-mer whose Hamming distance from the leafnode is $m$ (like $T/ACA(m = 0)$ or $T/AAA(m = 1)$ in (Fig. 2 (c)).

***Time Complexity of gkm-SVM:*** We also split operations of gkm-SVM into those indirectly (pre-processing) or directly (kernel-update) updating $\mathbf{N}_m$.

*Pre-processing:* To construct the trie, gkm-SVM iterates over every possible starting position for a $g$-mer. Given, there are $N$ sequences each of average length $l$, then there are approximately $Nl$ starting positions. Furthermore, each $g$-mer must be inserted into the trie ($g$ steps). Therefore, the time taken to construct the *trie* is $O(gNl)$. Besides, for each node (a unique $g$-mer) in the trie, the algorithm maintains a list of pointers that point to all other $g$-mers in the trie whose hamming distance to this node is $M$. Let the number of such $g$-mers be $\eta$ and total number of nodes are $ug$, then this operation costs $O(\eta ug)$.

*Kernel Update:* For each leafnode of the trie (total $u$ nodes), for each $g$-mer in its nodelist (assuming average size of nodelist is $\eta$), gkm-SVM uses the matching count among $g$-mers to update involved sequences' entries in $N_m$ (if Hamming distance between two $g$-mers is $m$). Therefore the time cost is $O(\eta u N^2)$ (at the worst case). Essentially $\eta$ represents on average the number of unique $g$-mers (in the trie) that are at a Hamming distance up to $M$ from the current leafnode. $\eta$ can be formulated as:

$$\eta = \min \left( u, \sum_{m=0}^{M=(g-k)} \binom{g}{m}(\Sigma - 1)^m \right) \sim \min \left( u, c_{gk}(\Sigma - 1)^M \right) \quad (10)$$

Fig. 3 shows that $\eta$ grows exponentially to $M$ until reaching its maximum $u$. The total complexity of time cost from gkm-SVM is thus $O(gNl + \eta ug + u\eta N^2)$.

***Comparing Time Complexity of GaKCo with gkm-SVM:*** Table 2 compares the asymptotic time cost of GaKCo with gkm-SVM. In gkm-SVM the term $O(\eta u N^2)$ dominates the overall time asymptotically. For GaKCo the term $O(c_{gk} z N^2)$ dominates the time cost asymptotically. For simplicity, we assume that $z = u$ even though $z \leq u$. Upon comparing $O(\eta \times u N^2)$ of gkm-SVM with $O(c_{gk} \times u N^2)$ of GaKCo, clearly the difference lies between the terms $\eta$ and $c_{gk}$.

In gkm-SVM, for a given $g$-mer, the number of all possible $g$-mers that are at a distance $M$ from it is $c_{gk}(\Sigma - 1)^M$. That is because $\binom{g}{M}$ positions can be substituted with $(\Sigma - 1)^M$ possible characters. This means $\eta$ grows exponentially with the number of allowed mismatches $M$. We show the trend of function $f = c_{gk}(\Sigma - 1)^M$ in Fig. 3 for three different applications - TF-DNA ($\Sigma = 4$), SCOP-protein ($\Sigma = 20$) and text ($\Sigma = 36$) when varying the values of $M$ for $g = 10$.

However, in real-world datasets, $\eta$ is upper bounded by the number of unique $g$-mers in a dataset: $u$. We show this by thresholding the curves in Fig. 3 at $u = 6 \times 10^4$, which is the average observed value of $u$ across multiple datasets. This means, two possible cases for comparing $\eta$ with $c_{gk}$:

- When $\eta \sim c_{gk}(\Sigma - 1)^M$: For cases whose dictionary size $\Sigma$ is small (e.g. 4), $c_{gk}(\Sigma - 1)^M$ is mostly smaller than $u$. Therefore $\eta$ will be close to $c_{gk}(\Sigma - 1)^M$. This indicates the costs of gkm-SVM grow with a speed proportional to $\Sigma^M$. In contrast, the term $c_{gk}$ of GaKCo is independent of the size $\Sigma$.
- When $\eta \sim u$: For cases whose $\Sigma$ is larger than 4, $c_{gk}(\Sigma - 1)^M$ gets larger than $u$ for $M \geq 4$. Therefore $\eta$ is approximately by $u$ (the number of unique $g$-mers in the trie built by gkm-SVM). The comparison between $u$ and $c_{gk}$ then depends on the specific application. The size $u$ depends on data, but normally grows fast for $M \geq 4$. For example, for one of the SCOP datasets, when $g = 10$, the count of unique $g$-mers $u = 6 \times 10^4$ at $M = 4$ (close to $u$ shown in Fig. 3). This means $\eta = 6 \times 10^6$ for gkm-SVM while for the same case $c_{gk} = 210$ for GaKCo. The former is approximately 300 times higher than GaKCo.

## 3 Experiments

### 3.1 Experimental Setup

***19 different sequence datasets:*** We perform 19 different classification tasks to evaluate the performance of GaKCo. These tasks belong to three categories: (1) Transcription Factor (TF) binding site prediction (DNA dataset), (2) Remote Protein Homology prediction (protein dataset), and (3) Character-based English text classification (text dataset). Table 3 summarizes of data statistics of all datasets we used. Details of these datasets and their associated applications are in the supplementary.

***Baselines:*** We compare the kernel calculation times and empirical performance of GaKCo with gkm-SVM [11]. We also compare GaKCo to the CNN implementation from [17] for all the datasets (results in supplementary).

Table 3. Details of datasets used for different prediction tasks. All tasks, except WebKB, are binary classification tasks. WebKB is a multi-class (4) classification dataset.

| Prediction Task | Repo | Datasets | Training | | Testing | | Sample properties | | |
|---|---|---|---|---|---|---|---|---|---|
| | | | Pos seq | Neg seq | Pos seq | Neg seq | $N$ | $\Sigma$ | Max($l$) |
| TF Binding Site (DNA) | ENCODE | CTCF | 1000 | 1000 | 1000 | 1000 | 4000 | 5 | 100 |
| | | EP300 | | | | | | | |
| | | JUND | | | | | | | |
| | | RAD21 | | | | | | | |
| | | SIN3A | | | | | | | |
| Remote Protein Homology (Protein) | SCOP | 1.1 | 1150 | 1189 | 8 | 1227 | 3574 | 20 | 905 |
| | | 1.34 | 866 | 1209 | 6 | 1231 | 3312 | | |
| | | 2.19 | 110 | 1235 | 9 | 1206 | 2560 | | |
| | | 2.31 | 1063 | 1235 | 8 | 1194 | 3500 | | |
| | | 2.1 | 4763 | 1229 | 120 | 950 | 7062 | | |
| | | 2.34 | 286 | 1215 | 6 | 1231 | 2738 | | |
| | | 2.41 | 192 | 1235 | 6 | 1213 | 2646 | | |
| | | 2.8 | 56 | 1185 | 8 | 1231 | 2480 | | |
| | | 3.19 | 922 | 1181 | 7 | 1231 | 3341 | | |
| | | 3.25 | 1187 | 1208 | 11 | 1231 | 3637 | | |
| | | 3.33 | 466 | 1214 | 7 | 1231 | 2918 | | |
| | | 3.50 | 105 | 1231 | 8 | 1205 | 2549 | | |
| Text Classification | Stanford Treebank | Sentiment | 3883 | 3579 | 877 | 878 | 9217 | 36 | 260 |
| | Dataset from [6] | WebKB | 335, 620, 744, 1083 | | 166, 306, 371, 538 | | 4163 | 36 | 14218 |

***Classification:*** After calculation, we input the $N \times N$ kernel matrix into an SVM classifier as an empirical feature map using a linear kernel in LIBLINEAR [10]. Here $N$ is the number of sequences in each dataset. For the multi-class classification of WebKB data, we use the multi-class version of LIBSVM [7].

***Model parameters:*** We vary the hyperparameters $g \in \{7, 8, 9, 10\}$ and $k \in \{1, 2, \ldots, g-1\}$ of both GaKCo and gkm-SVM. $M = (g - k)$ for all these cases. We also tune the hyperparameter $C \in \{0.01, 0.1, 1, 10, 100, 1000\}$ for the SVM. We present the results for the best $g$, $k$, and $C$ values based on the empirical performance metric.

***Evaluation Metrics:*** *Running time:* We compare the kernel calculation times of GaKCo and gkm-SVM in seconds. All run-time experiments were performed on an AMD Opteron™ Processor 6376 @ 2.30GHz with 250GB memory.

*Empirical performance:* We use the Area Under Curve (AUC) score (from the Receiver Operating Characteristic (ROC) curve) as our empirical evaluation metric for 18 binary classification tasks. We report the results of WebKB multi-class classification using micro-averaged F1 score.

### 3.2 Experimental Results

***GaKCo is as accurate as gkm-SVM:*** Fig. 1 (b) demonstrated that GaKCo achieves the same empirical performance as gkm-SVM across all 19 tasks (on AUC scores or F1-score). This is because GaKCo's gapped $k$-mer formulation is the same as gkm-SVM but with an improved (faster) implementation. Besides, in the supplementary, we also compare GaKCo's empirical performance with a state-of-the-art CNN model [17]. For 16/19 tasks, GaKCo outperforms the CNN model with an average of $\sim 20\%$ improvements.

***GaKCo scales better than gkm-SVM for larger dictionary size ($\Sigma$) and larger number of mismatches ($M$):*** Fig. 1(a) shows that GaKCo is faster than gkm-SVM for 16/19 tasks. The three tasks for which GaKCo cost similar time in kernel calculation as gkm-SVM are three DNA sequence prediction tasks. This is as expected since these tasks have a smaller dictionary ($\Sigma = 5$) and thus, for a small number of allowed mismatches ($M$) gkm-SVM gives comparable speed performance as GaKCo.

Fig. 4 shows the kernel calculation times of GaKCo versus gkm-SVM for the best-performing $g$ and varying $k \in \{1, 2, \ldots (g-1)\}$ for three binary classification datasets: (a) EP300 (DNA), (b) 1.34 (protein), and (c) Sentiment (text) respectively. We select these three datasets as they achieve the best AUC scores out of all 19 tasks (see supplementary). We fix $g$ and vary $k$ to show time performance for any number of allowed mismatches from 1 to $g - 1$. For GaKCo, the results are shown for both single-thread and the multi-thread implementations. We refer to the multi-thread implementation as GaKCo because that is our final code version. Our results show that GaKCo (single-thread) scales better than gkm-SVM for a large dictionary size ($\Sigma$) and a large number of mismatches ($M$). The final version of GaKCo (multi-thread) further improves the performance. Details for each dataset are as follows:

- DNA dataset ($\Sigma = 5$): In Fig. 4 (a), we plot the kernel calculation times for best $g = 10$ and varying $k$ with $M \in \{1, 2, \ldots 9\}$ for EP300 dataset. As expected, since the dictionary size of DNA dataset ($\Sigma$) is small, gkm-SVM performs fast kernel calculations for $M = (g - k) < 4$. However, for large $M \geq 4$, its kernel calculation time increases considerably compared to GaKCo. This result connects to Fig. 3 in Section 2, where our analysis showed that the *nodelist* size becomes closer to $u$ as $M$ increases, thus increasing the time cost.
- Protein dataset ($\Sigma = 20$): Fig. 4 (b), shows the kernel calculation times for best $g = 10$ and varying $k$ with $M = (g - k) \in \{1, 2, \ldots 9\}$ for 1.34 dataset. Since the dictionary size of protein dataset ($\Sigma$) is larger than DNA, gkm-SVM's kernel calculation time is worse than GaKCo even for smaller values of $M < 4$. This also connects to Fig. 3 where the size of *nodelist* $\sim u$ even for small $M$ for protein dataset, resulting in higher time cost. For best-performing parameters $g = 10, k = 1(M = 9)$, gkm-SVM takes 5 hours to calculate the kernel, while GaKCo uses less than 4 minutes.
- Text dataset ($\Sigma = 36$): Fig. 4 (c), shows the kernel calculation times for best $g = 8$ and varying $k$ with $M \in \{1, 2, \ldots 7\}$ for Sentiment dataset. For large $M \geq 4$, kernel calculation of gkm-SVM is slower as compared to GaKCo. One would expect that with large dictionary size ($\Sigma$) the performance difference will be same as that for protein dataset. However, unlike protein sequences, where the substitution of all 20 characters in a $g$-mer is roughly equally likely, text dataset has a more skewed underlying distribution. The chance of substituting some characters in a $g$-mer are higher than other characters for English text. For example, in a given $g$-mer "my nam", the last position is more likely to be occupied by 'e' than 'z'. Though the dictionary size is large here, the growth of the *nodelist* is restricted by the underlying distribution. While GaKCo's time performance is consistent across all three datasets, gkm-SVM's time performance varies due to the distribution properties.

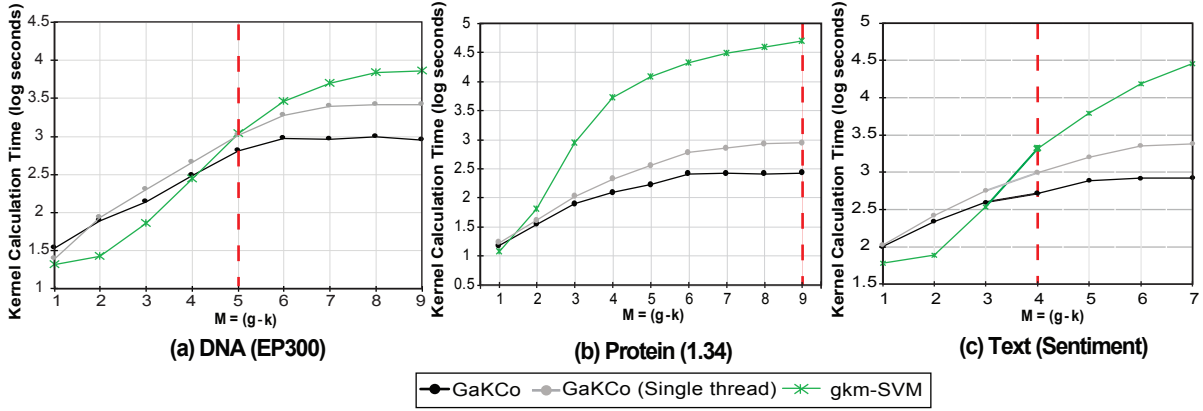

**Fig. 4.** Kernel calculation times (lower is better) for best $g$ and varying $k$ with $M = (g - k) = \{1, 2, \ldots g - 1\}$ hyperparameters for (a) EP300 (DNA, $\Sigma = 5$), (b) 1.34 (protein, $\Sigma = 20$), and (c) Sentiment (text, $\Sigma = 36$) datasets. The best performing hyperparameters ($g, k$ or $M = (g - k)$) are highlighted as red colored dashed lines. GaKCo (single thread) outperforms gkm-SVM for a large dictionary size ($\Sigma > 5$) and a large number of mismatches $M \geq 4$. The final GaKCo (multi-thread) implementation further improves the performance. For protein dataset (b) gkm-SVM takes $> 5$ hours to calculate the kernel, while GaKCo calculates it in 4 minutes.

According to our asymptotic analysis in Section 2, GaKCo should always be faster than gkm-SVM. However, in Fig. 4 we notice that for certain cases (e.g. for DNA when $M < 4$ in Fig. 4) GaKCo's is slower than gkm-SVM. This is because, in our analysis, we theoretically estimate the size of gkm-SVM's *nodelist*. In practice, we see that the actual *nodelist* size is smaller than our estimated for some cases. Among those cases for some gkm-SVM is faster than GaKCo. However, when with a larger value of $M(\geq 4)$ or a larger dictionary ($\Sigma > 5$), the nodelist size in practice matches our theoretical estimation; therefore, GaKCo always has lower kernel calculation time complexity than gkm-SVM for these cases.

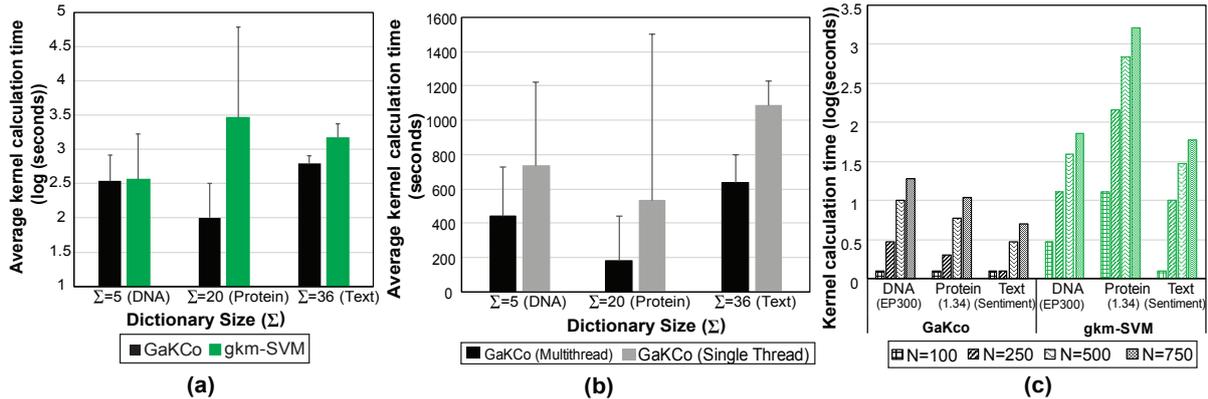

**Fig. 5.** Average kernel calculation times (lower is better) (a) for the best performing $(g, k)$ parameters for DNA ($\Sigma = 5$), protein ($\Sigma = 20$), and text ($\Sigma = 36$) datasets. gkm-SVM slows down considerably for protein and text datasets but GaKCo is consistently faster for all three datasets. (b) across DNA (5), protein (12) and text (2) datasets. Multi-thread GaKCo implementation improves the kernel calculation speed of the single-thread GaKCo by a factor of 2. (c) Kernel calculation times (lower is better) of GaKCo and gkm-SVM for best performing parameters $(g, k)$ for: EP300 (DNA), 1.34 (protein), and Sentiment (text) datasets. Length of the sequences for all three datasets is fixed to $l = 100$ and number of sequences are varied for $N \in \{100, 250, 500, 750\}$. With increasing number of sequences, the increase in kernel calculation time is more drastic for gkm-SVM than for GaKCo across all three datasets.

***GaKCo is independent of dictionary size ($\Sigma$):*** GaKCo's time complexity analysis (Section 2) shows that it is independent of the $\Sigma^M$ term, which controls the size of gkm-SVM's *nodelist*. In Fig. 5 (a), we plot the average kernel calculation times for the best performing $(g, k)$ parameters for DNA ($\Sigma = 5$), protein ($\Sigma = 20$), and text ($\Sigma = 36$) datasets respectively. The results validate our analysis. We find that gkm-SVM takes similar time as GaKCo to calculate the kernel for DNA dataset due to the small dictionary size. However, when the dictionary size increases for protein and text datasets, it slows down considerably. GaKCo, on the other hand, is consistently faster for all three datasets, despite the increase in dictionary size.

***GaKCo algorithm benefits from parallelization:*** As discussed earlier, the calculation of $\mathbf{C}_m$ (with $m \in \{0, 1, \ldots, M = (g - k)\}$) is an independent procedure in GaKCo's algorithm. This property makes GaKCo naturally parallelizable. We implement the final parallelized version of GaKCo by distributing calculation of each $\mathbf{C}_m$ on its thread. In Fig. 4 we see that the multi-threaded version of GaKCo performs faster than its single-threaded counterpart. Next, in Fig. 5(b), we plot the average kernel calculation times across DNA (5), protein (12) and text (2) datasets for both multi-thread and single thread implementations. Hence, we demonstrate that the improvement in speed by parallelization is consistent across all datasets.

***GaKCo scales better than gkm-SVM for increasing number of sequences ($N$):*** We now compare the kernel calculation times of GaKCo versus gkm-SVM for increasing number of sequences ($N$). In Fig. 5(c), we plot the kernel calculation times of GaKCo and gkm-SVM for best performing parameters $(g, k)$ for three binary classification datasets: EP300 (DNA), 1.34 (protein), and Sentiment (text). We select these three datasets as they provide the best AUC scores out of all 19 tasks (see supplementary). To show the effect of increasing $N \in \{100, 250, 500, 750\}$ on kernel calculation times, we fix the length of the sequences for all three datasets to $l = 100$. As expected, the time grows for both the algorithms with the increase in the number of sequences. However, this growth in time is more drastic for gkm-SVM than for GaKCo across all three datasets. Therefore, GaKCo is ideal for adaptive training since its kernel calculation time increases more gradually than gkm-SVM as new sequences are added. Besides, GaKCo's time improvement over the baseline is achieved with almost no added memory cost (see supplementary).

## 4   Conclusion

In this paper, we presented GaKCo, a fast and naturally parallelizable algorithm for gapped $k$-mer based string kernel calculation. The advantages of this work are:

- **Fast:** GaKCo is a novel combination of two efficient concepts: (1) reduced gapped $k$-mer feature space and (2) associative array based counting method, making it faster than the state-of-the-art gapped $k$-mer string kernel, while achieving same accuracy. (Fig. 1).
- GaKCo can **scale up** to larger values of $m$ and $\Sigma$. (Fig. 4 and Fig. 5(a))
- **Parallelizable:** GaKCo algorithm naturally leads to a parallelizable implementation (Fig. 4 and Fig. 5 (b))
- We have provided a detailed **theoretical analysis** comparing the asymptotic time complexity of GaKCo with gkm-SVM. This analysis, to the best of the authors' knowledge, has not been reported before (Section 2.3).

# Supplementaray Document for
# GaKCo: a Fast <u>Ga</u>pped *k*-mer string <u>K</u>ernel using <u>Co</u>unting

Ritambhara Singh, Arshdeep Sekhon, Jack Lanchantin,
Kamran Kowsari, Beilun Wang and Yanjun Qi

Department of Computer Science, University of Virginia (`yanjun@virginia.edu`)

## S:1 More Details of GaKCo Algorithm

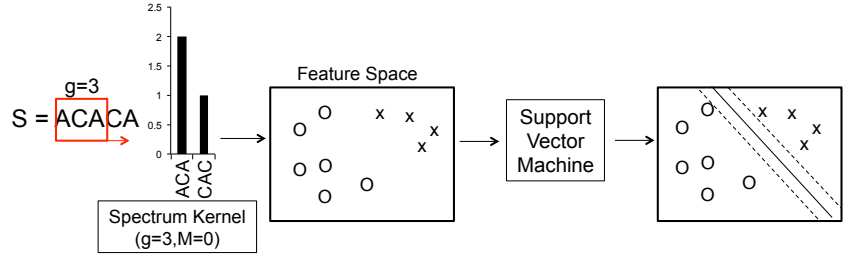

**Fig. S:1.** Overview of the String (Spectrum) Kernel + SVM classifier.

### S:1.1 Formal proof regarding Hamming Distance Property

Let hamming distance between strings $x$ and $y$ be $d(x,y)$. Assuming both $x$ and $y$ are composed of $n$ characters, then hamming distance is formally defined as [2]:

$$d(x,y) = \Sigma_{i=0}^{n} neq(x_i, y_i) \tag{S:1–1}$$

where, if $a$ and $b$ are two characters,

$$neq(a,b) = \begin{cases} 0, & \text{if } a = b \\ 1, & \text{otherwise} \end{cases} \tag{S:1–2}$$

**Property:** Given, there are two strings $x$ and $y$ (composed of $n$ characters each) and characters from $p$ positions are removed to obtain strings $x'$ and $y'$ with $(n-p)$ characters. If the hamming distance between $x'$ and $y'$, $d(x', y') = 0$ then the hamming distance between original $x$ and $y$, $d(x,y) \leq p$.

**Proof by example:**
Let $p = 2$. We first re-write Eq. S:1–1 as:

$$d(x,y) = \Sigma_{i=0}^{n-2} neq(x_i, y_i) + neq(x_{n-1}, y_{n-1}) + neq(x_n, y_n) \tag{S:1–3}$$

That is, we split the summation of $neq(.)$ function as summation of $neq(.)$ for $(n-2)$ characters plus the sum of $neq(.)$ for the $(n-1)^{th}$ and last $n^{th}$ character for $x$ and $y$.

The term $\Sigma_{i=0}^{n-2} neq(x_i, y_i)$ represents the hamming distance $d(x', y')$ for $p = 2$ positions removed. Therefore:

$$d(x,y) = d(x', y') + neq(x_{n-1}, y_{n-1}) + neq(x_n, y_n) \tag{S:1–4}$$

Now if $d(x', y') = 0$ then

$$d(x,y) = neq(x_{n-1}, y_{n-1}) + neq(x_n, y_n) \tag{S:1–5}$$

Based on Eq. S:1–2, $d(x,y) = \{(0+0), (0+1), (1+0), (1+1)\}$ as these are all the possible values of $neq(.)$ function.

Therefore, $d(x,y) \leq 2$ if $d(x', y') = 0$ where $x'$ and $y'$ are $x$ and $y$ (respectively) with characters removed from $p = 2$ positions.

### S:1.2 Justification of GaKCo's Sort and Count Method

A core piece of the GaKCo's kernel computation is counting the observed $g$-mers in the strings for which the kernel value is being computed. The final implementation of our algorithm uses a sorting-based counting method, but we did consider a hashing approach as well. There are straightforward time complexity justifications for choosing sorting over hashing, which we explain in this section.

A hash table, treated as an associative array, could easily be used to count instances of a $g$-mer. Given a $g$-mer, which consists of a $g$-length token and a reference to the original string number, we may write a simple hash function that executes in $\Theta(g)$ time (as we ought to consider every character in the string for a well-distributed hash). Also, given that the total number of strings is $N$ of average length $l$, then the total number of $g$-mers is $\sim Nl$. If we accept the "typical-case" runtime of insertion into a hash table, which is $\Theta(1)$, then to count every $g$-mer we must perform at least $\Theta(g \cdot Nl)$ steps: for each of the total $Nl$ $g$-mers, we do $g$ work to hash, insert, and update the associated value.

At first consideration, a sorting-based approach would seem to be strictly worse, as any swapping sort would take $\Theta(g \cdot (Nl) \lg(Nl))$ time. However, using a non-swapping sort, in our case, gives us $\Theta(g \cdot Nl)$ time, which is the same as we derived for the above hashing method. However, the sorting requires nearly exactly $g \cdot Nl$ steps, while the hashing approach needs more steps to resolve any possible collisions. To confirm our theoretical justification, we implemented hashing approach and found that our sorting method was, indeed, faster than hashing.

## S:2 Connecting to Previous Studies

**String Kernels** Aside from the spectrum kernel [18] and gapped $k$-mer kernel [11], a few other notable string kernels include (but are not limited to): (1) *($k, m$)-Mismatch Kernel.* This kernel calculates the dot product of contiguous $k$-mer counts with $m$ mismatches allowed. The cumulative matching statistic idea was first proposed by [15] for this kernel. [1] (2) *Substring Kernel.* It measures the similarity between sequences based on common co-occurrence of exact matching subpatterns (e.g., substrings) [32]. (3) *Profile Kernel.* This method uses the notion of similarity based on a probabilistic model (e.g. profile) [14]. (4) *Cluster Kernel.* The "sequence neighborhood" kernel or "cluster" kernel [8] is a semi-supervised extension of the string kernel. It replaces every sequence with a set of "similar" (neighboring) sequences and obtains a new representation. Then, it averages over the representations of these contiguous sequences found in the unlabeled data using a sequence similarity measure.

All string kernels calculate the feature representation $\phi(\cdot)$ using the counts of $k$-mer occurrence. Thus, in the following paragraph we will briefly discuss notable methods that count the occurrence of $k$-mers (mostly in the bioinformatics literature).

**$k$-mer counting methods** $k$-mer (or in our case, $g$-mer) counting is the method by which we determine the number of matching or unique $k$-mers (or $g$-mers) in any text or pattern. Tools handling large text datasets need to filter out these unique $k$-mers (or $g$-mers) to reduce the processing or counting time. GaKCo uses a 'sort and count' method for calculating the number of matching $g$-mers to compute the mismatch profile. This is a widely used method that lists all the $g$-mers, sorts them lexicographically and counts all the consecutive matching entries while skipping the unique $g$-mers. It has been used previously in tools used for genome assembly [23], discovery of motifs (or most common fixed length patterns) [26], and string kernel calculation [15].

**BioSequence Classification with Deep Learning** In recent years, deep learning models have become popular in the bioinformatics community, owing to their ability to extract meaningful representations from large labeled datasets (e.g., with sample size $\sim$30,000 sequences). For example, Qi et al. [25] used a deep multi-layer perceptron (MLP) architecture with multitask learning to perform sequence-based protein structure prediction. Zhou et al. [34] created a generative stochastic network to predict secondary structure on the same data as used by Qi et al. [25]. Recently, Lin et al. [20] outperformed all the state-of-the-art works for protein property prediction task by using a deep convolutional neural network architecture. Later, Alipanahi et al. [1] applied a convolutional neural network model for predicting sequence specificity of DNA and RNA-binding proteins as well as generating motifs, or consensus patterns, from the features that were learned by their model. Lanchantin et al. [17] proposed a deep convolutional/highway MLP framework for the same task and demonstrated improved performance. In the field of natural language processing, multiple works like [30] have used deep learning models for document [33] or sentiment [29] classification.

---

[1] Because [15] uses all possible $k$-mers built from the dictionary with $m$ mismatches as the feature space, the authors [15] need to precompute a complex weight matrix to incorporate all possible $k$-mers with $m$ mismatches.

## S:3 Details about the Datasets

### S:3.1 Benchmark Tasks of Sequence Classification

*DNA and Protein Sequence Classification* Studying DNA and Protein sequences gives us deeper insight into the biological processes that can, in turn, help us understand cell development and diseases. Two major tasks essential in the field are Transcription Factor Binding Site (TFBS) Prediction and Remote Protein Homology Prediction.

Transcription factors (TFs) are regulatory proteins that bind to functional sites of DNA to control the regulation of genes. Each different TF binds to specific locations (or sites) on a genomic sequence to regulate cell machinery. Owing to the development of chromatin immunoprecipitation and massively parallel DNA sequencing (ChIP-seq) technologies [24], maps of genome-wide binding sites are currently available for multiple TFs across different organisms. Because ChIP-seq experiments are slow and expensive, computational methods to identify TFBSs accurately are essential for understanding cell regulation.

Remote Protein Homology Prediction, i.e. classification of protein sequences according to their biological function or structure, plays a significant role in drug development. Protein sequences that are a part of the same protein superfamily are evolutionarily related and functionally and structurally relevant to each other [4]. Protein sequences with feature patterns showing high homology are classified into the same superfamily. Once assigned a family, the properties of the protein can be easily narrowed down by analyzing only the superfamily to which it belongs.

Researchers have formulated both these tasks as classification tasks, where knowing a DNA or protein sequence, we would like to classify it as a binding site or non-binding site for TF prediction and belonging or not belonging to a protein family for homology prediction respectively.

*Text Classification* Text classification incorporates multiple tasks like assigning subject categories or topics to documents, spam detection, language detection, sentiment analysis, etc. Generally, given a document and a fixed number of classes, the classification model has to predict the class that is most relevant to that document. Several recent studies have discovered that character-based representation provides straightforward and powerful models for relation extraction [28], sentiment classification [33], and transition based parsing [5]. Lodhi et. al. [21] first used string kernels with character level features for text categorization. However, their kernel computation used dynamic programming which was computationally intensive. Over recent years, more efficient string kernel methods have been devised [18, 14–16, 11]. Therefore, we use simple character-based text input for document and sentiment classification tasks.

### S:3.2 19 Datasets used in Evaluations

*ENCODE DNA Sequences:* Transcription factors (TFs) are regulatory proteins that bind to functional sites of DNA to control the regulation of genes. Each different TF binds to specific locations (or sites) on a genomic sequence to regulate cell machinery. Maps of genome-wide binding sites are currently available for multiple TFs for human genome via the ENCODE [9] database. These "maps" mark the positions of the TF binding sites. We select 100 basepair sequences overlapping the binding sites as positive sequences and randomly select non-binding sequences from the human genome as negative sequences. We perform this selection for five different transcription factors (CTCF, EP300, JUND, RAD21, and SIN3A) from the K562 (leukemia) cell type, resulting in five different prediction tasks. For these tasks, we use a dictionary size of 5 ($\Sigma = 5$); four nucleotide symbols — A, T, C, G — and a "unknown" character 'N' for nucleotides that were not read by the sequencing machines. Therefore, the dictionary is {A,T,C,G,N}.

*SCOP Protein Sequences:* Remote Protein Homology Prediction, i.e. classification of protein sequences according to their biological function or structure, plays a significant role in drug development. Protein sequences that are a part of the same protein superfamily are evolutionarily related and functionally and structurally relevant to each other [4]. The SCOP domain database consists of protein domains, no two of which have 90% or more residual identity [13]. It is hierarchically divided into folds, superfamilies, and finally families. We use 12 sets of samples (listed in Supplementary Table) and select positive test sequences (for each sample) from 1 protein family of a particular superfamily. We obtain the positive training sequences from remaining families in that superfamily. We select negative training and test sequences from non-overlapping folds outside the positive sequence fold. We use the dictionary size of 20 ($\Sigma = 20$) as there are 20 amino acid symbols that make up a protein sequence. Therefore the dictionary is {A,C,D,E,F,G,H,I,K,L,M,N,P,Q,R,S,T,V,W,Y}.

*WebKB and Sentiment Classification Datasets:* Text classification incorporates multiple tasks like assigning subject categories or topics to documents, spam detection, language detection, sentiment analysis, etc. Several recent studies have discovered that character-based representation provides straightforward and powerful models for relation extraction [28], sentiment classification [33], and transition based parsing [5]. We downloaded the processed WebKB datasets (removed stop/short words, stemming, etc.) from [6]. This task is a multi-class classification task of webpages with four classes: project, course, faculty, and student. For the sentiment analysis experiments, we used the Stanford sentiment treebank dataset [29]. This dataset provides a score for each sentence between 0 and 1 with $[0, 0.4]$ being negative sentiment and $[0.6, 1.0]$ being positive. We combined the validation set in the original treebank dataset with the training set. We use the dictionary size of 36 ($\Sigma = 36$) since we use character-based input. The dictionary includes all the alphabets [A-Z] and numbers [0-9].

## S:4  Empirical Performance of GaKCo versus Neural Networks

Recently, Deep Neural Networks (NNs) have provided state-of-the-art performances for various sequence classification tasks like analyzing DNA [1, 17], proteins [25, 34], and natural language [33, 29] sequences. Despite their superior performance in accuracy and speed (e.g. through GPUs and mini-batches) such NN systems usually require a significant number of training samples. This requirement can be unfeasible for many datasets, especially in the medical research domains. Here, the number of training sequences per experiment can be as low as tens or hundreds due to cost and time constraints. We compare GaKCo's empirical performance with a state-of-the-art deep convolutional neural network (CNN) model [17]. On datasets with few training samples, GaKCo achieves an average accuracy improvement of 20% over the CNN model (see Fig. S:2) making it an appealing tool when the training samples are scarce. Besides, GaKCo includes only two hyperparameters ($g$ and $k$) for tuning [2]. This feature is desirable when comparing with NN systems for which figuring out the optimal network model and hyperparameters can be a daunting task.

More concretely, we compare GaKCo's empirical performance with a state-of-the-art CNN model from [17]. Fig. S:2 (b) shows the differences in AUC Scores (or micro-averaged F1-score for Web-KB) of GaKCo and CNN [17]. For 16/19 tasks, GaKCo outperforms the CNN model with an average of $\sim 20\%$ accuracy. This result can be explained by the fact that CNNs trained with a small number of samples (1000-10,000 sequences) often exhibit unstable behavior in performance.

For three datasets - SIN3A (DNA), 1.1 (protein), and Web-KB (text), we observe that the empirical performance of GaKCo and CNN is similar. Therefore, we further explore these datasets in Fig. S:2(c). Here, we plot the AUC scores or micro-averaged F1 scores (Web-KB) for varying number of training sample ($N = \{100, 250, 500$ and $750\}$ sequences). We randomly select these samples from the training set and use the original test set of the respective datasets. The results are averaged over three runs of the experiment. Our aim is to find the threshold (number of training samples) for which CNN gives a lower performance to GaKCo for these three datasets. Fig. S:2(c) presents the averaged AUC scores or micro-averaged F1 score (Web-KB). We see that the threshold for which CNN gives a lower performance to GaKCo is 750 sequences in the training set. We also observe that the variance in performance is high for NN (represented by error bars) across the three runs.

## S:5  Other Experiments

### S:5.1  Time profiles for different functions of GaKCo

Table S:1 shows the GaKCo average time profiled for sorting vs. count and update function when calculating cumulative mismatch profile for EP300 DNA dataset ($m = 7$).

---

[2] There is also one $C$ parameter for tuning SVM training (while using linear kernel)

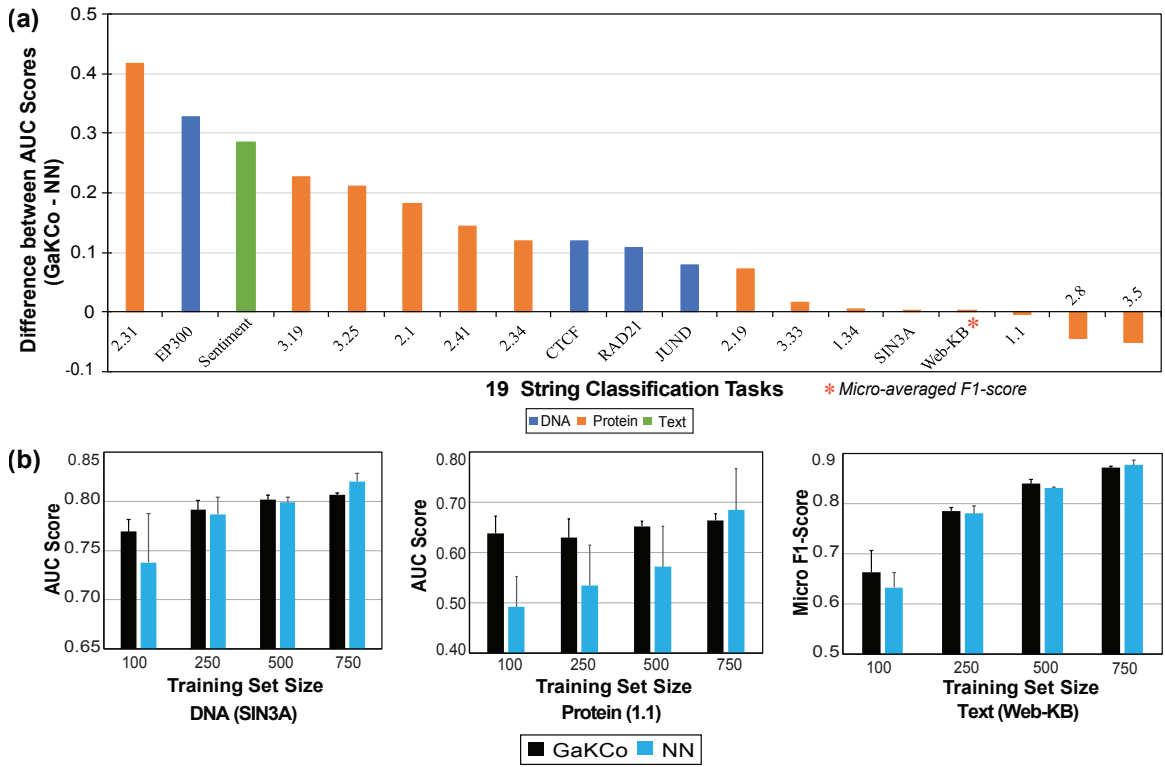

**Fig. S:2.** (a) Kernel calculation time (X-axis) and the memory usage (Y-axis) (lower is better) for both GaKCo and gkm-SVM for all 19 classification tasks. For 17/19 tasks, GaKCo's memory usage is lesser or comparable to gkm-SVM with lower kernel calculation time. Therefore, it is both time and memory efficient. (b) Differences in AUC Scores (or micro-averaged F1-score for Web-KB) between GaKCo and state-of-the-art CNN model [17]. For 16/19 tasks, GaKCo outperforms CNN with an average of ∼ 20% accuracy. (c)Averaged AUC scores, across 3 runs, for SIN3A (DNA) and 1.1 (protein), and micro-averaged F1 scores for Web-KB (text) while varying number of sequences ($N = \{100, 250, 500,$ and $750\}$). For a threshold value of 750 sequences in the training set, CNN achieves lower empirical performance to GaKCo.

**Table S:1.** The GaKCo sorting time vs Count and Update time averaged over all iterations for EP300 DNA dataset (g=10 and k=3).

| Algorithmic Step | Average Time (in seconds) |
|---|---|
| Sort | 0.059 |
| Count and Update | 2.26 |

### S:5.2 AUC scores for the best performing parameters

*Different handling of dictionaries* Table S:2 summarizes the AUC scores for all datasets. The current gkm-SVM implementation [11] while reading the input ignores an unknown character. GaKCo maps it to another 'UNK' character. For example, the dataset may contain an extra 'X' character, which is not a part of the dictionary. To ensure consistent empirical performance between GaKCo and gkm-SVM, the user will have to add the extra 'X' character to the dictionary of gkm-SVM.

### S:5.3 Running Time Results of GaKCo with one level of parallelization versus GaKCo with two levels of parallelization

As explained earlier, our GaKCo implementation utilizes the parallelizability of GaKCo over iterations over $m$ mismatches. It is possible to parallelize GaKCo on another level as the calculation of the cumulative mismatch profile $C_i$ is also independent for the $\binom{g}{i}$ iterations i.e. the Step 4 in Fig 2 (Main

---
[4] * indicates memory issues

**Table S:2.** Summary of GaKCo-SVM, gkm-SVM and CNN-AUC scores for all datasets. For Web-KB we report the micro-averaged F1-Score since it is a multi classification task with four classes: student, faculty, project and course.

| Prediction Task | Sample properties | | | Best Parameters | | | AUC | | |
|---|---|---|---|---|---|---|---|---|---|
| Datasets | $N$ | $\Sigma$ | Max($l$) | g | k | c | GaKCO-AUC | gkm-SVM-AUC | NN-AUC |
| 1.1 | 3574 | | | 7 | 5 | 0.01 | 0.7453 | 0.7448 | 0.7484 |
| 1.34 | 3312 | | | 10 | 1 | 0.1 | 0.9903 | 0.9903 | 0.9858 |
| 2.19 | 2560 | | | 7 | 1 | 100 | 0.8951 | 0.8951 | 0.822 |
| 2.31 | 3500 | | | 10 | 7 | 10 | 0.9484 | 0.9497 | 0.5317 |
| 2.1 | 7062 | | | 10 | 3 | 10 | 0.979 | 0.9895 | 0.7970 |
| 2.34 | 2738 | 20 | 905 | 7 | 6 | 0.01 | 0.8664 | 0.8660 | 0.7477 |
| 2.41 | 2646 | | | 10 | 6 | 0.01 | 0.7925 | 0.7925 | 0.6484 |
| 2.8 | 2480 | | | 10 | 1 | 10 | 0.6367 | 0.6367 | 0.6801 |
| 3.19 | 3341 | | | 8 | 1 | 0.1 | 0.9326 | 0.9326 | 0.7050 |
| 3.25 | 3637 | | | 10 | 8 | 1 | 0.7967 | 0.7962 | 0.5848 |
| 3.33 | 2918 | | | 10 | 5 | 0.01 | 0.9018 | 0.9018 | 0.8843 |
| 3.50 | 2549 | | | 10 | 7 | 0.01 | 0.7768 | 0.7772 | 0.8265 |
| CTCF | | | | 10 | 5 | 1 | 0.902 | 0.902 | 0.7834 |
| EP300 | | | | 10 | 5 | 1 | 0.942 | 0.942 | 0.6138 |
| JUND | 4000 | 5 | 100 | 10 | 7 | 1 | 0.91 | 0.91 | 0.8317 |
| RAD21 | | | | 10 | 5 | 1 | 0.901 | 0.901 | 0.7937 |
| SIN3A | | | | 10 | 7 | 1 | 0.834 | 0.834 | 0.8309 |
| Sentiment | 9217 | 36 | 260 | 8 | 4 | 1 | 0.8154 | 0.81 | 0.5303 |
| WebKB (F1-score) | 4163 | 36 | 14218 | 8 | 5 | 1 | 0.9153 | 0.9116 | 0.9147 |

**Table S:3.** GaKCo-Parallel (Single Level Parallelization) vs GaKCo-Parallel+ (two levels of parallelization) for WebKB dataset

| Levels of Parallelization | Time (seconds) | Memory(GB) |
|---|---|---|
| GaKCo-Parallel | 751 | 1.63 |
| GaKCo-Parallel+ | 58 | 24.44 |

**Table S:4.** GaKCo-Parallel(Single Level Parallelization) vs GaKCo-Parallel+ (two levels of parallelization). The time is in seconds[4]

| Dataset | GaKCo-Parallel | GaKCo-Parallel+ |
|---|---|---|
| 1.1 | 31 | 13 |
| 1.34 | 266 | 63 |
| 2.19 | 79 | 28 |
| 2.31 | 120 | 17 |
| 2.1 | 974 | * |
| 2.34 | 10 | 7 |
| 2.41 | 90 | 19 |
| 2.8 | 184 | 43 |
| 3.19 | 175 | 35 |
| 3.25 | 54 | 15 |
| 3.33 | 151 | 32 |
| 3.50 | 58 | 10 |
| WebKB | 751 | 58 |
| Sentiment | 522 | 137 |

Text) can be done independently over all $\binom{g}{i}$ positions. We also performed similar experiments for a two-level parallelization implementation of GaKCo. As summarized in Table S:3 and Table S:4, the kernel calculation time decreases manifold, but this speed-up comes at the cost of increased memory usage. If memory is not a constraint, our GaKCo-Parallel+ implementation can be used to further speed up kernel calculation.

### S:5.4 GaKCo is both time and memory efficient:

Fig. S:3 shows points for the kernel calculation time (X-axis) versus the memory usage (Y-axis) for both GaKCo and gkm-SVM for all 19 classification tasks. We observe that most of these points representing GaKCo lie in the **lower-left quadrant** indicating that it is both time and memory efficient. For 17/19 tasks, its memory usage is lesser or comparable to gkm-SVM with faster kernel calculation time. Therefore, GaKCo's time improvement over the baseline is achieved with almost no added memory cost.

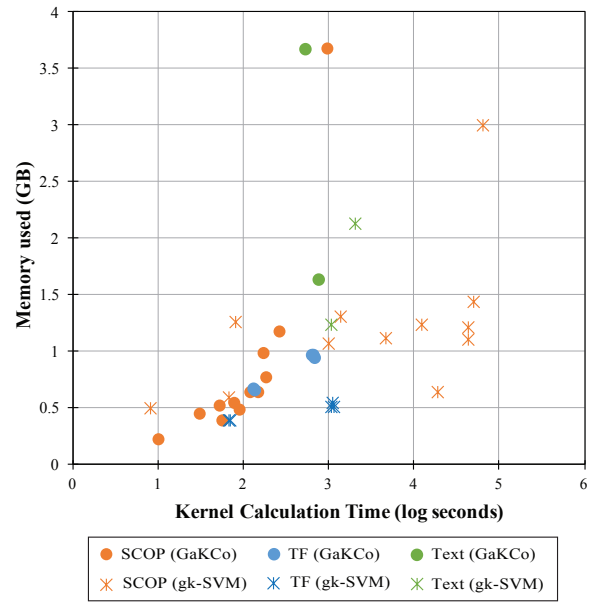

**Fig. S:3.** Kernel calculation time (X-axis) and the memory usage (Y-axis) (lower is better) for both GaKCo and gkm-SVM for all 19 classification tasks. For 17/19 tasks, GaKCo's memory usage is lesser or comparable to gkm-SVM with lower kernel calculation time. Therefore, it is both time and memory efficient.